%% file: root.tex
\documentclass[letterpaper, 10 pt, conference]{ieeeconf} 

\IEEEoverridecommandlockouts                        

\usepackage{graphics}
\usepackage{amsmath} 
\usepackage{amssymb}
\usepackage{cite}
\usepackage{tikz}
\usepackage{amsmath}
\usepackage{tabularx}
\usepackage{array}
\usepackage{booktabs}
\usepackage{bm}
\usepackage{algorithm}
\usepackage{algorithmic}
\usepackage{bbm}
\usepackage{hyperref}
\usepackage{units}
\usepackage[caption=false,font=normalsize,labelfont=sf,textfont=sf]{subfig}
\usepackage{caption}
\usepackage{tablefootnote}

\definecolor{green}{RGB}{3,200,15}

\title{\LARGE \bf
Autonomous Ground Navigation in Highly Constrained Spaces: \\Lessons learned from The 3rd BARN Challenge at ICRA 2024
}
\author{\textbf{Competition Organizers}: Xuesu Xiao$^{1}$, Zifan Xu$^{2}$, Aniket Datar$^{1}$, Garrett Warnell$^{2, 3}$, and Peter Stone$^{2, 4}$, 
\\\textbf{Winning Teams}: Joshua Julian Damanik$^{5}$, Jaewon Jung$^{5}$, Chala Adane Deresa$^{5}$, \\ Than Duc Huy$^{6}$, Chen Jinyu$^{6}$, Chen Yichen$^{6}$, Joshua Adrian Cahyono$^{6}$, \\  Jingda Wu$^{7}$, Longfei Mo$^{7}$, Mingyang Lv$^{7}$, Bowen Lan$^{7}$, Qingyang Meng$^{7}$, Weizhi Tao$^{7}$, and Li Cheng$^{7}$
\thanks{$^{1}$George Mason University
$^{2}$The University of Texas at Austin
$^{3}$Army Research Laboratory
$^{4}$Sony AI 
$^{5}$Korea Advanced Institute of Science \& Technology (KAIST)
$^{6}$Nanyang Technological University (NTU)
$^{7}$The Hong Kong polytechnic University (PolyU)
}
}
\begin{document}

\maketitle
\thispagestyle{empty}
\pagestyle{empty}

\input{content/abstract.tex}
\input{content/barn.tex}
\input{content/simulation.tex}
\input{content/physical.tex}

\input{content/teams.tex}
\input{content/discussions.tex}

\input{content/conclusions.tex}

\bibliographystyle{IEEEtran}
\bibliography{IEEEabrv,references}
\end{document}

%% file: content/abstract.tex
\begin{abstract}

The 3rd BARN (Benchmark Autonomous Robot Navigation) Challenge took place at the 2024 IEEE International Conference on Robotics and Automation (ICRA 2024) in Yokohama, Japan and continued to evaluate the performance of state-of-the-art autonomous ground navigation systems in highly constrained environments. 
Similar to the trend in The 1st and 2nd BARN Challenge at ICRA 2022 and 2023 in Philadelphia (North America) and London (Europe), The 3rd BARN Challenge in Yokohama (Asia) became more regional, i.e., mostly Asian teams participated. The size of the competition has slightly shrunk (six simulation teams, four of which were invited to the physical competition). 
The competition results, compared to last two years, suggest that the field has adopted new machine learning approaches while at the same time slightly converged to a few common practices. However, the regional nature of the physical participants suggests a challenge to promote wider participation all over the world and provide more resources to travel to the venue. 
In this article, we discuss the challenge, the approaches used by the three winning teams, and lessons learned to direct future research and competitions. 

\end{abstract}

%% file: content/barn.tex
\section{The 3rd BARN Challenge Overview}
\label{sec::challenge}

The 3rd BARN  (Benchmark Autonomous Robot Navigation) Challenge~\cite{the_3rd_barn_challenge} took place as a conference competition at the 2024 IEEE International Conference on Robotics and Automation (ICRA 2024) in Yokohama, Japan. As a continuation of The 1st and 2nd BARN Challenge at ICRA 2022 and 2023 in Philadelphia and London respectively, the 3rd challenge aimed to evaluate the capability of state-of-the-art navigation systems to move robots through static, highly-constrained obstacle courses, an \emph{ostensibly} simple problem even for many experienced robotics researchers, but in fact, as the results from the first two competitions suggested, a problem far away from being solved~\cite{xiao2022autonomous, xiao2023autonomous}. 

Each team needed to develop an entire navigation software stack for a standardized and provided mobile robot, i.e., a Clearpath Jackal~\cite{clearpath_jackal} with a 2D 270\textdegree-field-of-view Hokuyo LiDAR for perception and a differential drive system with $2\textrm{m/s}$ maximal speed for actuation.
The developed navigation software stack needed to autonomously drive the robot from a given starting location through a dense obstacle field and to a given goal without any collisions with obstacles or any human interventions.
The team whose system could best accomplish this task within the least amount of time would win the competition.
The 3rd BARN Challenge had two phases: a qualifying phase evaluated in simulation, and a final phase evaluated in three physical obstacle courses.
The qualifying phase took place before the ICRA 2024 conference using the BARN dataset~\cite{perille2020benchmarking} (with the recent addition of DynaBARN~\cite{nair2022dynabarn}), which is composed of 300 obstacle courses in Gazebo simulation randomly generated by cellular automata.
The top four teams from the simulation phase were then invited to compete in three different physical obstacle courses set up by the organizers at ICRA 2024 in the PACIFICO Yokohama conference center.

In this article, we report on the simulation qualifier and physical finals of The 3rd BARN Challenge at ICRA 2024, present the approaches used by the top three teams, discuss lessons learned from the challenge compared against The 1st and 2nd BARN Challenge at ICRA 2022 and 2023, and point out future research directions to solve the problem of autonomous ground navigation in highly constrained spaces. 

%% file: content/simulation.tex
\section{Simulation Qualifier}
\label{sec::simulation}
\begin{figure*}[t]
    \centering
    \includegraphics[width=2\columnwidth]{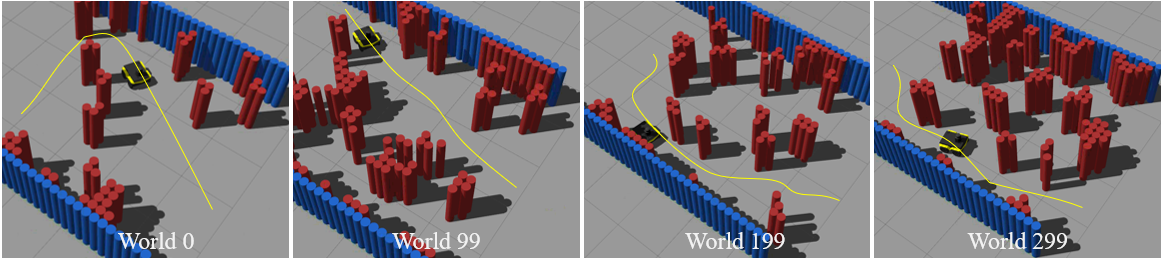}
    \caption{Four example BARN environments in the Gazebo simulator (ordered by ascending relative difficulty level).}.
    \label{fig::barn_worlds}
\end{figure*}

The simulation qualifier of The 3rd BARN Challenge started on January 1\textsuperscript{st}, 2024. 
The qualifier used the BARN dataset~\cite{perille2020benchmarking}, which consists of 300 $5\textrm{m}\times5\textrm{m}$ obstacle environments randomly generated by cellular automata (see examples in Fig.~\ref{fig::barn_worlds}), each with a predefined start and goal. 
These obstacle environments range from relatively open spaces, where the robot barely needs to turn, to highly dense fields, where the robot needs to squeeze between obstacles with minimal clearance.
The BARN environments are open to the public, and were intended to be used by the participating teams to develop their navigation stack.
Another 50 unseen environments, which are not available to the public, were generated to evaluate the teams' systems.
A random BARN environment generator was also provided to the teams so that they could generate their own unseen test environments.\footnote{\url{https://github.com/dperille/jackal-map-creation}}

In addition to the 300 BARN environments, six baseline approaches were also provided for the participants' reference, ranging from classical sampling-based~\cite{fox1997dynamic} and optimization-based navigation systems~\cite{quinlan1993elastic}, to end-to-end machine learning methods~\cite{xubenchmarking, wang2021agile}, and hybrid approaches~\cite{xu2021applr}.
All baselines were implementations of different local planners used in conjunction with Dijkstra's search as the global planner in the ROS \texttt{move\_base} navigation stack~\cite{rosmovebase}.
Additionally, the winning teams' navigation stacks from the last two competitions were also open sourced~\cite{the_3rd_barn_challenge}. 
To facilitate participation, a training pipeline capable of running the standardized Jackal robot in the Gazebo simulator with ROS Noetic (in Ubuntu 20.04), with the option of being containerized in Docker or Singularity containers for fast and standardized setup and evaluation, was also provided.\footnote{\url{https://github.com/Daffan/ros_jackal}}

\subsection{Rules}
Each participating team was required to submit their developed navigation system as a (collection of) launchable ROS node(s).
The challenge utilized a standardized evaluation pipeline\footnote{\url{https://github.com/Daffan/nav-competition-icra2022}} to run each team's navigation system and compute a standardized performance metric that considers navigation success rate (collision or not reaching the goal counts as failure), actual traversal time, and environment difficulty (measured by optimal traversal time).
Specially, the score $s$ for navigating each environment $i$ was computed as
\[
s_i = 1^{\textrm{success}}_i \times \frac{\textrm{OT}_i}{\textrm{clip}(\textrm{AT}_i, 2\textrm{OT}_i, 8\textrm{OT}_i)} \; ,
\]
where the indicator function $1_\textrm{success}$ evaluates to $1$ if the robot reaches the navigation goal without any collisions, and evaluates to $0$ otherwise.
$\textrm{AT}$ denotes the actual traversal time, while $\textrm{OT}$ denotes the optimal traversal time, as an indicator of the environment difficulty and measured by the shortest traversal time assuming the robot always travels at its maximal speed ($2\textrm{m/s}$):
\[
\textrm{OT}_i = \frac{\textrm{Path Length}_i}{\textrm{Maximal Speed}}.
\]
The Path Length is provided by the BARN dataset based on Dijkstra's search from the given start to goal.
The $\textrm{clip}$ function clips $\textrm{AT}$ within 2OT and 8OT in order to assure navigating extremely quickly or slowly in easy or difficult environments respectively won't disproportionally scale the score. 
Notice that the lower bound 2OT was reduced from the previous 4OT used in the last two challenges, considering the performance upper bound, 0.25, has been closely approached by multiple teams. In The 3rd BARN Challenge, the upper bound has been increased to 0.5 to encourage faster navigation speed.  
The overall score of each team is the score averaged over all 50 unseen test BARN environments, with 10 trials in each environment. Higher scores indicate better navigation performance.
The six baselines score between $0.1656$ and $0.4354$~\cite{the_3rd_barn_challenge}.

\subsection{Results}
The simulation qualifier started on January 1\textsuperscript{st}, 2024 and lasted through a soft submission deadline (April 1\textsuperscript{st}, 2024) and a hard submission deadline (May 1\textsuperscript{st}, 2024). Submitting by the soft deadline will guarantee an invitation to the final physical competition given good navigation performance in simulation and leave sufficient time for invited participants to make travel arrangements to Yokohama. The hard deadline is to encourage broader participation, but final physical competition eligibility will depend on the available capacity and travel arrangement made beforehand.  
In total, six teams, five from Asia and one from Europe, submitted their navigation systems.
The performance of each submission was evaluated by the standard evaluation pipeline. The results are shown in Tab.~\ref{tab::sim_results} with the baselines shown in the fourth column as a reference. 

\begin{table}[h]
  \caption{Simulation Results.}
  \label{tab::sim_results}
  \centering
  \footnotesize
  \begin{tabular}{cccc}
  \toprule
  Rank. & Team & Score & Baseline \\
  \midrule
  1 & LiCS-KI & 0.4762\\
  2 & AIMS & 0.4723 & LfLH~\cite{wang2021agile}, e2e~\cite{xubenchmarking} \\
  3 & EIT-NUS & 0.3795 & APPLR-DWA~\cite{xu2021applr}, E-Band~\cite{quinlan1993elastic}\\
  4 & MLDA\_EEE & 0.2476 & (Fast \& Default) DWA~\cite{fox1997dynamic}\\
  5 & Tartu Team & NA\\
  6 & CCWSS & NA\\
  \bottomrule
  \end{tabular}
\end{table}

The top two simulation teams, LiCS-KI from Korea Advanced Institute of Science and Technology (KAIST) and AIMS from The Hong Kong Polytechnic University (HKPU) outperformed all last year's winning teams (KUL+FM, INVENTEC, and University of Almeria). However, there is still a gap between the new performance upper bound (0.5) and the top performance (0.4762).  
The top four teams, LiCS-KI, AIMS, EIT-NUS from Eastern Institute of Technology, Ningbo, China, and MLDA\_EEE from Nanyang Technological University (NTU) were invited to the physical finals at ICRA 2024. The top simulation (and also final winning) team, LiCS-KI, was the only team that submitted after the soft deadline but before the hard deadline. 

%% file: content/physical.tex
\section{Physical Finals}
\label{sec::physical}
\begin{figure}[ht]
    \centering
    \includegraphics[width=1\columnwidth]{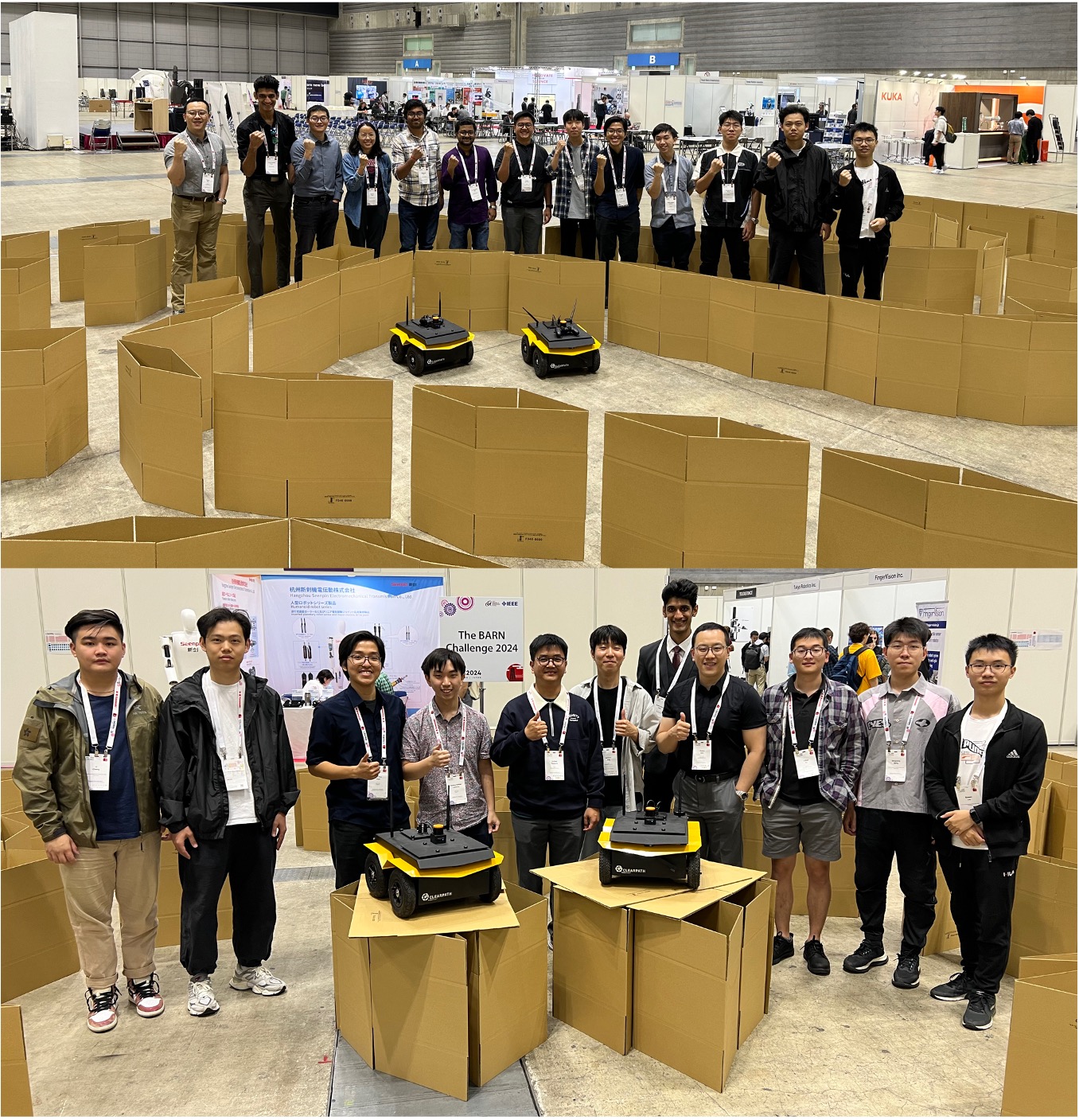}
    \caption{Final physical competition participants and organizers at The 3rd BARN Challenge in Yokohama, Japan. }
    \label{fig::barn_at_icra}
\end{figure}

The physical finals took place at ICRA 2024 in the PACIFICO Yokohama conference center on May 15\textsuperscript{th} and May 16\textsuperscript{st}, 2024 (Fig.~\ref{fig::barn_at_icra}).
Two physical Jackal robots with the same sensors and actuators were provided by the competition sponsor, Clearpath Robotics. 

\subsection{Rules}
Physical obstacle courses were set up using 120 cardboard boxes in the conference center.
The organizers used the same guidelines to set up three obstacle courses as in the first two BARN challenges, i.e., all courses aimed at testing a navigation system's local planning and therefore had an obvious passage but with minimal clearance (a few centimeters around the robot) when traversing this passage. Considering that KUL+FM finished all three physical obstacle courses in The 2nd BARN Challenge, the organizers intentionally increased the difficulty this year, i.e., introducing sharper turns and smaller clearances. 

The organizers also used the same competition rules agreed upon by all the physical competition participants: Each team has 20 minutes to set up their navigation system after each obstacle course was constructed. After the 20-minute set-up time, each team had the opportunity to run five timed trials (after notifying the organizers to be timed) within another 20-minute period. The fastest three out of the five timed trials were counted, and the team that had the most successful trials (reaching the goal without any collision) would be the winner. 
In the case of a tie, the team with the fastest average traversal time would be declared the winner. 

\subsection{Results}
The four teams' navigation performance is shown in Tab.~\ref{tab::physical_results}. 
Due to the intentionally increased navigation difficulty, the teams struggled more on obstacle avoidance, similar to The 1st BARN Challenge, and focused less on increasing speed, as the teams did during The 2nd BARN Challenge. 
The detailed results of all five timed trials (in seconds, only the top three were counted in the final score) are listed in the last three columns of Tab.~\ref{tab::physical_results}, where ``X'' indicates failure. 

The winner, LiCS-KI, successfully and quickly finished all ten trials in the first two courses, but failed all five trials in the 3rd course, the extremely difficult one.  MLDA\_EEE also completely failed in the last, most difficult course, but succeeded in three and two trials in the first two courses. AIMS was able to slowly but successfully finish three trials in the last course, but did not perform well in the first two, especially the 2nd course, possibly due to a bug caused by sensor dimension mismatch. As a result, LiCS-KI won the competition by the most successful trials (6/9), while the tie between MLDA\_EEE and AIMS was broken by the average traversal time (79s vs. 109s). 

\begin{table*}[ht]
  \caption{Physical Results.}
  \label{tab::physical_results}
  \centering
  \small
  \begin{tabular}{ccccccc}
  \toprule
  Rank. & Team & Success/Total & Average Time & Course 1 & Course 2 & Course 3 \\
  \midrule
  1 & LiCS-KI & 6/9 & 30/35/NA & 32/31/32/27/30 & 37/37/40/29/32 & X/X/X/X/X\\
  2 & MLDA\_EEE & 5/9 (79) & 72/89/NA & 68/X/77/X/70 & X/X/X/85/93 & X/X/X/X/X\\
  3 & AIMS & 5/9 (109) & 90/NA/121 & 92/88/X/X/X & X/X/X/X/X & 119/118/126/X/X\\
  4 & EIT-NUS & 0/9 & NA/NA/NA & X/X/X/X/X & X/X/X/X/X & X/X/X/X/X \\
  \bottomrule
  \end{tabular}
\end{table*}

%% file: content/teams.tex
\section{Top Three Teams and Approaches}
\label{sec::teams}
In this section, we report the approaches used by the three winning teams.

\subsection{LiCS-KI (KAIST)}

The LiCS-KI team from KAIST introduced an end-to-end local navigation method for indoor navigation and deployed their Learned-imitation on Cluttered Space (LiCS) framework \cite{damanik2024lics}. The main innovation is the use of a transformer-based network trained using Behavior Cloning (BC) with robust expert demonstrations under controlled noise. This method enables the robot to navigate robustly and rapidly through highly cluttered spaces. Additionally, a safety check layer is added to ensure safe navigation in untrained environments, particularly during real-world challenges.

\begin{figure}[ht]
    \centering
    \includegraphics[width=1\columnwidth]{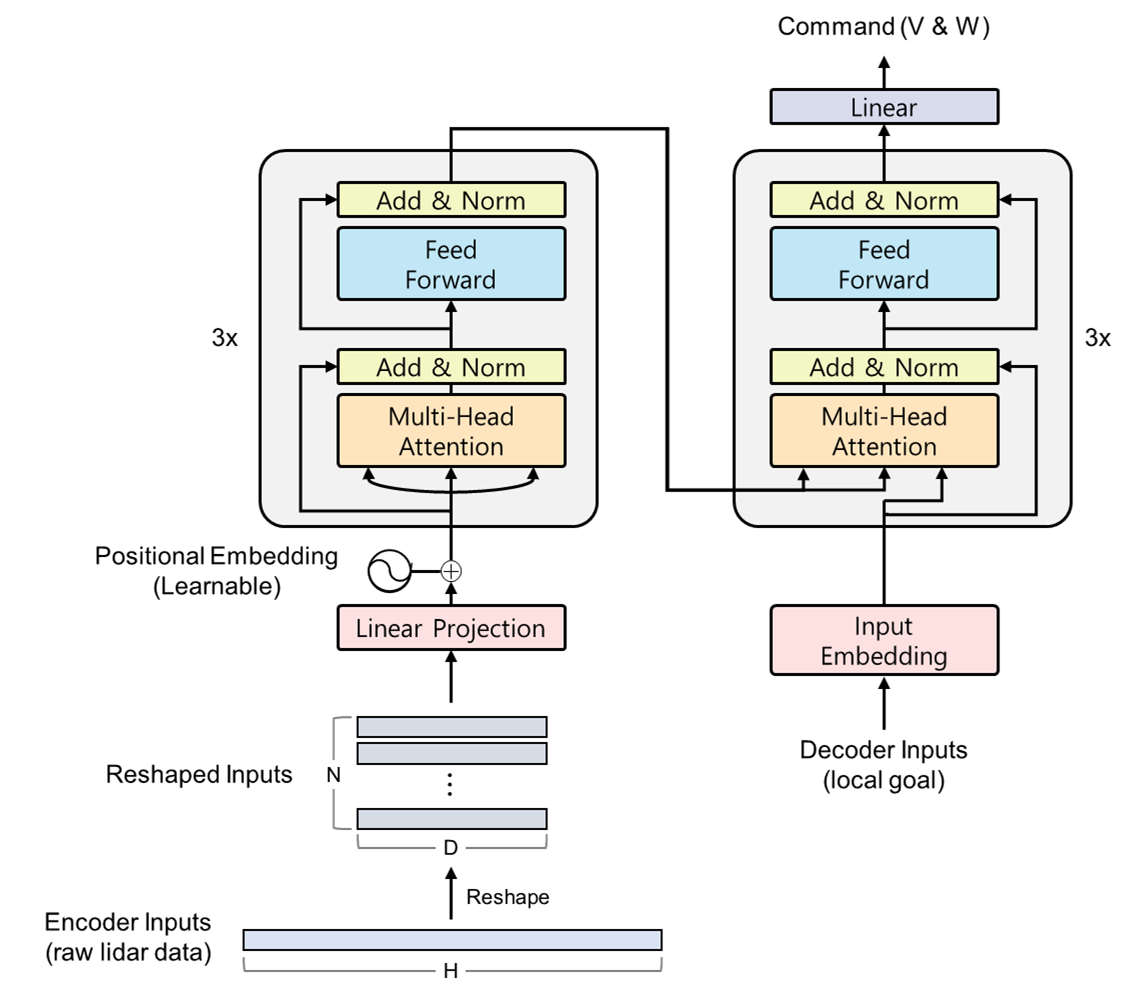}
    \caption{Transformer-based neural network used in LiCS \cite{damanik2024lics}.}
    \label{fig:lics_network}
\end{figure}

\subsubsection{\textbf{Neural network}}

The neural network used by LiCS consists of a Transformer encoder and decoder, as depicted in Fig. \ref{fig:lics_network}. The encoder employs a Vision Transformer (ViT) \cite{dosovitskiy2021image} model with class token omitted, while the decoder is a standard transformer decoder without positional embedding and masked multi-head attention. During the encoding process, the LiDAR input is segmented into $N$ patches, projected through a linear network, and concatenated with learnable position embedding. The decoder processes the encoded LiDAR data alongside the normalized local goal, provided by the global planner, to predict the optimal linear and angular velocities ($v$ and $\omega$).

\subsubsection{\textbf{Behavior Cloning (BC)}}

The proposed network is trained using BC to replicate the expert algorithm from the previous year's winning team, KUL+FM. To address the inherent performance issues possessed by BC \cite{ross2010efcient}, a Gaussian noise $\mathcal{N}(0, \sigma^2)$ is injected to the control inputs during expert demonstrations. This noise augmentation allows the demonstrations to cover a variety of states for training, enhancing the policy network's robustness \cite{green1986persistence, laskey2017dart, ke2021grasping}.

\subsubsection{\textbf{Safety check layer}}

\begin{figure}[ht]
\centering
\subfloat[Linear]{\includegraphics[height=1.5in]{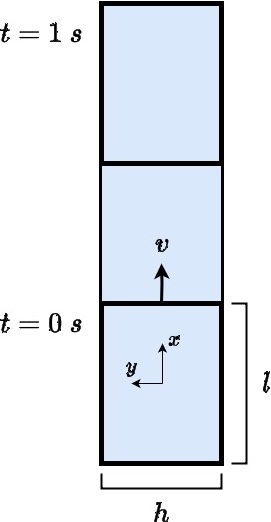}%
\label{fig:lics_safety_linear}}
\hfil
\subfloat[Radial]{\includegraphics[height=1.5in]{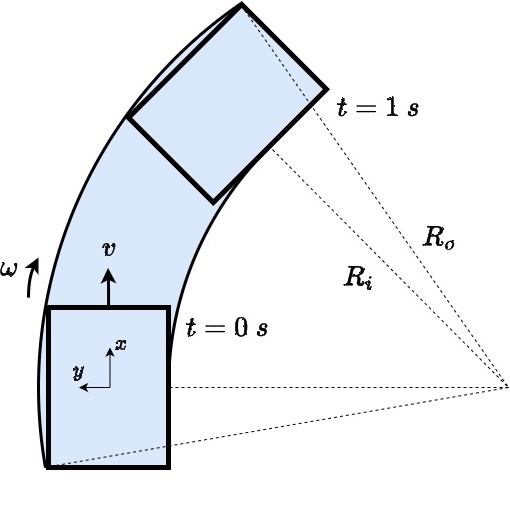}%
\label{fig:lics_safety_radial}}
\caption{Safety zone illustration for the safety check layer of LiCS during linear and radial movement.}
\label{fig:lics_safety}
\end{figure}

The safety check layer uses geometric calculations based on combined LiDAR and costmap data to enhance model safety. For linear motion ($|v| > 0,\ \omega = 0$), the robot travels along its x-axis. Safety is assessed by ensuring no obstacles are within a predefined rectangular safety zone extending from the robot's front (Fig. \ref{fig:lics_safety_linear}). In radial motion scenarios ($|v| > 0,\ |\omega| > 0$), where the robot follows a circular trajectory, the safety check involves ensuring no obstacles are present within two polygons that represent the robot’s footprint at the start and end of a movement interval, connected by arcs defining the robot's outer and inner turning radii (Fig. \ref{fig:lics_safety_radial}). Imminent collisions detected from this safety check layer trigger recovery actions, including speed reduction, in-place rotation, and backward movement.   

\subsubsection{\textbf{Implementation}}

The training dataset was collected by recording simulations of the KUL+FM approach with injected Gaussian noise ($\sigma=0.25$) across various scenarios. The network, consisting of three layers each in the ViT encoder and transformer decoder, was trained in a supervised manner using the combined dataset over 100 epochs. In both simulated and real-world challenges, an A* algorithm was used as the global planner with dynamic obstacle inflation parameters that are adjusted based on velocity, $r_\textrm{infl}=r_\textrm{min} + (v/v_\textrm{max}) \times (r_\textrm{max}-r_\textrm{min})$. The safety layer was implemented solely in the real-world challenge, as the obstacle courses differed from the simulation environments. \texttt{hector\_mapping} SLAM was also used during the real-world challenge to improve localization accuracy.

\subsection{MLDA\_EEE (NTU)}

Team MLDA\_EEE tackled the challenge using a classical approach with Model Predictive Control (MPC) with mode switching logic for different scenarios. The various modes all use MPC formulation with different initialization and constraints to bias the solver towards a feasible solution.

\subsubsection{\textbf{Formulation}}

The optimization variables of the MPC problem include the robot's coordinates and headings as the state variables, $\mathbf x = [x,y,\theta]$, and the velocity and acceleration of both left and right wheels as the control, $\mathbf u = [vr, vl, ar, al]$.
The MPC minimizes the objective function $J$ over the horizon of $N$ steps, subject to the constraints of wheeled differential-drive model, the current odometry readings, and other additional constraints, to make the robot follow a reference trajectory: 
$$\arg \min_{\mathbf{x,u}} \sum_{k=0}^{N-1} J\mathbf {(x,u)}.$$

The objective function includes (1) error to the reference trajectory taken from the global planner, (2) error reference velocity, and (3) acceleration: 
$$J = w_v(\vert v_k \vert - v_ref)^2$$
$$+ w_x[(x_k - x_{ref_k})^2 + (y_k - y_{ref_k})^2]$$
$$ + w_a(a_k - a_{k+1})^2.$$

\subsubsection{\textbf{Implementation}}

\begin{figure}[ht]
    \centering
    \includegraphics[width=1\columnwidth]{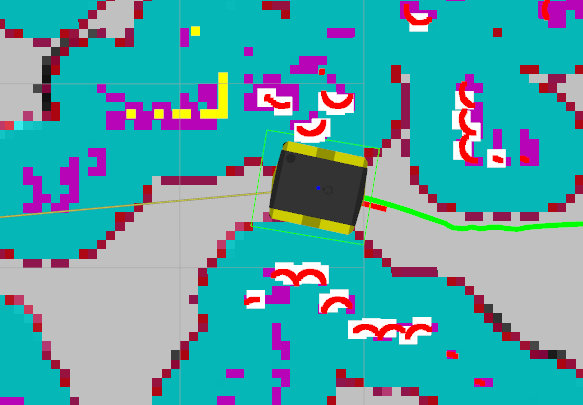}
    \caption{Rviz visualization of obstacles in the MPC. White squares: obstacle coordinates sampled from raw LiDAR scan. Yellow squares: blind spot obstacle coordinates obtained from \texttt{local\_costmap}. }
    \label{fig::mlda01}
\end{figure}

The global trajectory is given by the global planner from ROS \texttt{move\_base}  package. The \texttt{map\_server} is updated using \texttt{hector\_mapping} to increase the reliability of the costmap. To reduce the computation time in the MPC, we minimized the number of obstacles considered in the optimization process. A ROS node is used to sample the raw LiDAR scan every 15 points. The \texttt{local\_costmap} occupancy grid is used to obtain obstacles in the blind spot of the LiDAR, similar to INVENTEC \cite{xiao2023autonomous}. These are published as point clouds with $(x,y)$ coordinates used in the MPC as shown in Fig. \ref{fig::mlda01}. The local plan is obtained from optimizing the MPC problem using the non-linear solver CasADi \cite{Andersson2019}.

\subsubsection{\textbf{Behavior of Different Modes}}

Different modes have different MPC parameters such as the weights of the objective terms, control limits, and additional constraints on the reference global trajectory to allow safe maneuver near obstacles. These parameters are also fine-tuned in the physical runs.

\begin{figure}[t]
    \centering
    \includegraphics[width=1\columnwidth]{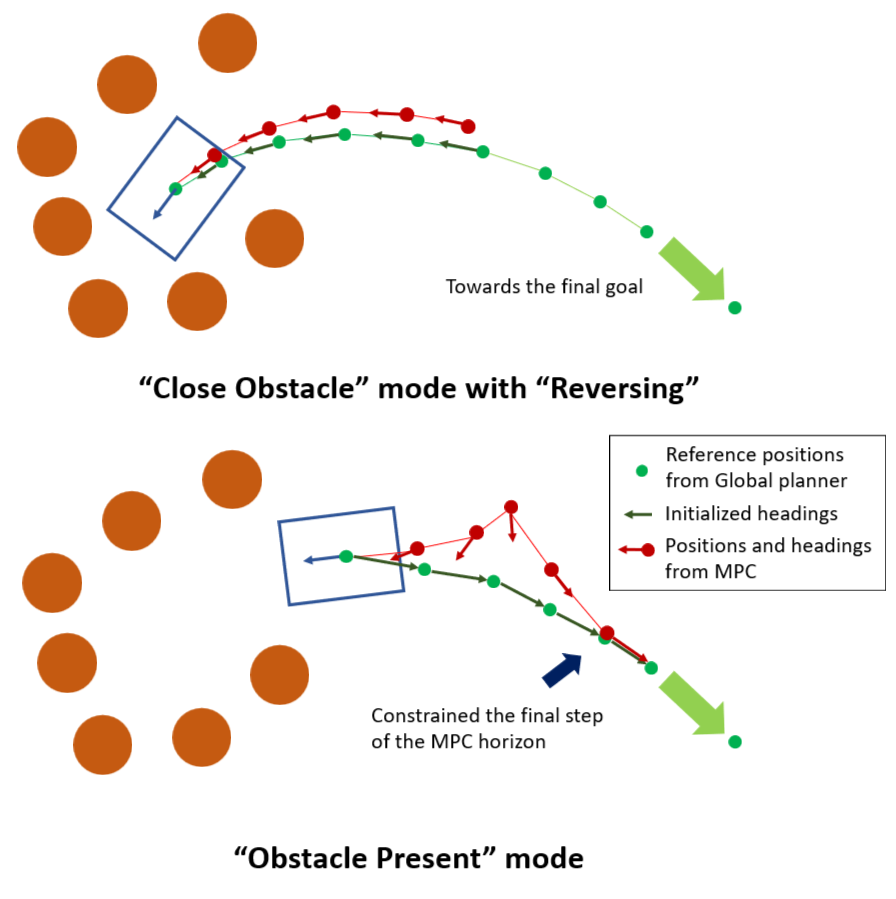}
    \caption{Different modes with different initialization and limits. ``Close obstacle'' with ``reversing'' mode has the headings pointing away from the goal to allow safe backtracking (top); ``Obstacle present'' mode has normal heading initialization and constraints allowing the robot to rotate to the correct heading (bottom).}
    \label{fig::mlda02}
\end{figure}

The various modes include (1) ``Safe'': the robot has high velocity, (2) ``Obstacle Present'': obstacle is detected 1m away from the robot, and (3) ``Close Obstacle'': obstacle is detected 0.5m away. Within the ``Close Obstacle'' mode, there is a ``Reversing'' mode which is triggered when the heading along the reference trajectory is more than $90^\circ$ from the current heading (Fig.~\ref{fig::mlda02} top). This happens when the robot encounters a dead-end and the global plan suggests a new trajectory. In this mode, the optimization variables are initialized such that the heading points away from the goal, priming the optimal solution to result in the robot backtracking, instead of performing a sudden rotation, to prevent collision with nearby obstacles. When the robot backtracks to a safe space, the ``Obstacle Present'' parameters and the constraints on the final MPC horizon step allow the robot to regain the correct heading towards the final goal (Fig.~\ref{fig::mlda02} bottom). 

\subsection{AIMS (PolyU)}

To effectively address the highly constrained environments in The BARN Challenge, the AIMS team developed a local planner utilizing advanced dynamic-window-based methods. This approach ensures collision-free navigation in narrow pathways by discretizing the sampling space into geometric elements for rapid determination in sensor space. It also incorporates backward sampling to assist the vehicle in adjusting its pose and extricating itself from tight situations. Given the competition rules prohibiting pre-mapping, the strategy focuses on real-time path planning rather than relying on real-time mapping and localization. This means the vehicle must continuously explore unknown areas while in motion, with the global plan constantly adjusted as exploration progresses. The algorithm is designed to continuously adapt to environmental changes and respond quickly to maintain collision-free navigation. The local planner operates without the need for global environmental data, allowing the vehicle to navigate challenging courses safely and efficiently, even without comprehensive maps and detailed localization information.

\subsubsection{\textbf{Efficient Geometry-Based Obstacle Detection}}

The method involves sampling potential trajectories with varying curvatures and applying geometric constraints to rapidly identify potential collision points among these predicted paths. This approach facilitates the selection of the optimal collision-free trajectory. Drawing inspiration from last year's winning team (KUL+FM), the local planner is integrated directly with the sensor data, thereby bypassing potential inaccuracies in the costmap and accelerating obstacle detection.

Specifically, the anticipated driving area is discretized into rectangles and triangles, arranged by proximity. By scanning these shapes for LiDAR-detected points to identify obstacles, it quickly determines whether the sampled areas are collision-free. Before this scanning process, a crucial step involves filtering the LiDAR points within each geometric shape based on their distance and angle relative to the sensor, significantly reducing the search space required for each geometric assessment. By leveraging direct sensor integration and geometric analysis, this enhanced method ensures real-time adjustments and precise obstacle detection, making the navigation solution robust against highly constrained environments.

\subsubsection{\textbf{Additional Backwards Sampling}}

A further enhancement to the traditional sampling-based method is the implementation of sampling during both forward and backward driving. This backward sampling design assists the vehicle in effectively adjusting its pose to extricate itself when direct forward movement is not possible. 

To be specific, both forward and backward driving involve sampling the potential trajectory curvatures of the vehicle. The difference lies in the judgment logic for backward sampling, which shifts from selecting samples that are closer to the local goal to selecting states that have more viable forward sampling points. This means that for each backward pose sampled, a second round of forward sampling is performed to find out those poses that have more feasible forward driving paths. A backward sampling result with more feasible forward paths typically indicates a superior vehicle pose in highly constrained environments, enabling better handling of complex situations when the vehicle reaches such a pose. 

%% file: content/discussions.tex
\section{Discussion}
\label{sec::discussions}

We discuss new findings and lessons from The 3rd BARN Challenge, not only from the technical perspective, but also from the competition organization side.

\subsection{First physical competition win of end-to-end learning }
In The 1st and 2nd BARN Challenges, the winning teams of the physical competition used classical approaches (UT AMRL and KUL+FM). However, the winning team this year, LiCS-KI, adopted an end-to-end imitation learning approach, which is the first physical win by end-to-end learning~\cite{xiao2022motion}. One interesting fact is that the expert used to provide demonstration data is a classical approach used by last year's winning team, KUL+FM, and LiCS-KI also added Gaussian noise to perturb the model input in order to achieve robustness, a classic data augmentation technique. Assisted by a Transformer architecture and safety check layer, LiCS-KI's approach outperformed its expert demonstrator, KUL+FM, in the simulation qualifier. It is worth noting that KUL+FM did not participate in the physical competition this year, so it is unclear whether the imitator can outperform the demonstrator in the physical runs. 

\subsection{First usage of Transformers in the challenge}
LiCS-KI is the first team in The BARN Challenge which used a Transformer architecture as the main local planner, compared to classical neural architectures used in the past years. The power of Transformers is one potential reason of LiCS-KI's win in both the simulation qualifier and physical finals, along with the data augmentation technique and safety check layer. The success of the Transformer architecture suggests the potential of better neural architecture for robot navigation tasks, not only to address visual inputs~\cite{nazeri2024vanp, kahn2021badgr, voilaharesh, sikand2022visual}, off-road conditions~\cite{datar2024terrain, datar2023learning, liang2024dtg, liang2023mtg, datar2023toward, xiao2021learning, karnan2022vi, pokhrel2024cahsor}, social contexts~\cite{raj2024rethinking, song2024socially, panigrahi2023study, mirsky2021prevention, francis2023principles, mavrogiannis2021core, scand, nguyen2023toward, xiao2022learning, hart2020using, pirk2022protocol}, kinodynamic constraints~\cite{das2023motion}, or multi-robot navigation~\cite{long2018towards, chen2017decentralized, park2023learning}, but also for purely geometric obstacle avoidance~\cite{xiao2021toward, xiao2021agile, wang2021agile, liu2021lifelong, ghani2024dyna}. The revolutionary success of Transformers on computer vision and natural language processing tasks may also inspire future navigation research. 

\subsection{Successful sim-to-real transfer of learning algorithms}
Similar to The 2nd BARN Challenge~\cite{xiao2023autonomous}, the 3rd year of the competition did not exhibit a significant sim-to-real gap. The first place winner of both simulation and physical course challenges, the LiCS-KI team, utilized a learning-based algorithm. Not only winning in terms of success rate during the physical finals, the team performed with the fastest average traversal time. This result highlights a small sim-to-real performance gap. It also suggests that learning-based models, particularly those trained in simulated environments, are becoming increasingly effective at handling the unpredictable nature of real-world settings when combined with the strategic use of imitation learning, specifically through behavior cloning, coupled with advanced data augmentation and neural architecture. This approach also contrasts with the more commonly used reinforcement learning~\cite{kahn2021badgr, voilaharesh, xu2023benchmarking, xu2021machine, xu2022learning} in past competitions. By employing imitation learning~\cite{pan2020imitation, xiao2021learning, karnan2022vi, atreya2022high, datar2023toward, datar2023learning}, the team was able to quickly deploy behaviors mimicking or even surpassing expert demonstrations, reducing the need for the trial-and-error learning phases typical of reinforcement learning. Additionally, the use of controlled noise during training helped the algorithm account for unforeseen variables and disturbances encountered. 
 
\subsection{Strong connections to previous years}
While last year's 2nd place winner, INVENTEC, based their approach on the strongest baseline, LfLH~\cite{wang2021agile} along the Learning from Hallucination (LfH)~\cite{xiao2021toward, xiao2021agile, wang2021agile, ghani2024dyna} line of work, the approaches developed by this year's teams started to show strong connections to previous years' methods. LiCS-KI used the approach by last year's KUL+FM as expert to generate demonstration data, while AIMS also leveraged KUL+FM's idea of local planning directly in the sensor space, instead of using costmaps, which are susceptible to inaccuracies. Along with the point in the following subsection, the community has started to form a few common practices to address the problem of navigation in highly constrained spaces, which also have real-world implications when deploying autonomous mobile robots in natural obstacle-occupied spaces. 

\subsection{Importance of a hybrid paradigm}
All teams adopted a hybrid paradigm in terms of a finite-state-machine setup, which requires different components to address different situations in the obstacle courses, especially safety checking of the actions produced by a main planner, differently parameterized MPC planners, and specifically designed reversing motions to back up the robot from undesirable scenarios. Such a pragmatic practice suggests that a single stand-alone approach that is able to address all variety of obstacle configurations all together is still out of our reach. Even for the end-to-end learning by LiCS-KI, a separate safety check layer is still required during hardware implementation. Most teams also specifically designed reversing or backtracking behaviors to address situations where the robot got stuck. However, more complex systems may introduce extra complications at the same time, e.g., proper parameter tuning for each component and appropriate transition conditions between different components.

\subsection{Tie-Breaking by Average Time for 2nd and 3rd Place}
Qualitatively speaking, this year's physical obstacle courses were slightly more difficult than last year's. Unfortunately, no team could finish all nine physical trials. LiCS-KI outperformed MLDA\_EEE and AIMS by one more successful trial, while MLDA\_EEE and AIMS were tied in terms of success rate. The tie was broken by average time. However, it is worth noting that due to the different length and difficulty of the three physical obstacle courses, it is difficult to make an absolutely fair comparison using average time of successful trials to break the tie: AIMS succeeded in three trials in the longest 3rd obstacle course, while didn't finish one single trial in the shorter 2nd obstacle course. MLDA\_EEE's performance was the opposite, which presents an advantage. Such a situation increases AIMS's average traversal time compared to MLDA\_EEE, causing the rank of 2nd and 3rd place.  

\subsection{More financial support for participation is needed}
One unfortunate fact about The 3rd BARN Challenge is that all four teams that participated in the physical finals are Asian teams. Considering that ICRA 2024 took place in Yokohama, Japan, not many teams from places far away from Japan  submitted their navigation stack to participate in the competition. The regional participation of the competition is not ideal to evaluate the entire field's progress and compare the performance of top teams all over the world. The organizers will try to reach out to more potential sponsors to provide more financial support to invite participants to travel from other continents. Another potential solution is to provide remote participation options, which was attempted last year. However, the need of fine-tuning the navigation systems for real-world deployment and to fit to every different obstacle course makes it impractical for the organizers to run the remote participants' systems and achieve reasonable performance out-of-the-box. How to remove the reliance on extensive system tuning is still an open question for robust obstacle avoidance in a variety of real-world scenarios~\cite{xiao2020appld,  wang2021appli, wang2021apple, xu2021applr, ma2021navtuner, xiao2022appl}.

%% file: content/conclusions.tex
\section{Future Plans}
\label{sec::conclusion}

Based on the first three year's BARN challenges, the organizers plan to make the following changes in the next BARN challenge in ICRA 2025. First, dynamic obstacles will be introduced to the currently static obstacle courses~\cite{nair2022dynabarn, ghani2024dyna}. For the first competition with dynamic obstacles, the organizers will allow collisions with dynamic obstacles and only add a penalty, whereas collisions with static obstacles will still be counted as a total failure. The addition of dynamic obstacles will stress-test the robustness of obstacle avoidance and also make the competition more interesting to watch. Second, to further encourage the teams to reduce the need of on-site fine-tuning, the organizers also plan to add a few ``cold trials'' at the beginning of each obstacle course: All teams will be required to directly navigate through each obstacle course without any fine-tuning of the system first. Successful cold trials will be rewarded by bonus points, before the teams are allowed to fine-tune their systems and start their regular trials. The organizers also plan on reducing the allowed time to fine-tune the system to discourage extensive dependence on manual trial and error before autonomous navigation. 